\documentclass[journal,comsoc]{IEEEtran}
\usepackage{amsmath,amsfonts}
\usepackage{algorithmic}
\usepackage{array}
\usepackage[caption=false,font=normalsize,labelfont=sf,textfont=sf]{subfig}
\usepackage{textcomp}
\usepackage{stfloats}
\usepackage{url}
\usepackage{verbatim}
\usepackage{graphicx,color}
\def\BibTeX{{\rm B\kern-.05em{\sc i\kern-.025em b}\kern-.08em
    T\kern-.1667em\lower.7ex\hbox{E}\kern-.125emX}}
\usepackage{balance}

\usepackage{optidef}
\usepackage{siunitx}
\usepackage{nicefrac}
\usepackage{multirow}
\usepackage{algorithm2e}
\DeclareMathOperator*{\argmax}{argmax}
\DeclareMathOperator*{\argmin}{argmin}
\newcommand{\Bernoulli}{\text{Bern}}
\RestyleAlgo{ruled}

\begin{document}
\title{A General Framework for Scalable UE-AP Association in User-Centric Cell-Free Massive MIMO based on Recurrent Neural Networks}

\author{Giovanni Di Gennaro, Amedeo Buonanno, \IEEEmembership{Senior Member, IEEE}, Gianmarco Romano, \\Stefano Buzzi, \IEEEmembership{Senior Member, IEEE}, and Francesco A.N. Palmieri, \IEEEmembership{Member, IEEE}
\thanks{A preliminary version of this paper was presented at the 2024 {\em IEEE International Workshop on Signal Processing Advances in Wireless Communications (SPAWC)} \cite{GDG_SPAWC2024}.}
\thanks{This paper was partly supported by the European Union under the Italian National Recovery and Resilience Plan (NRRP) of NextGenerationEU, partnership on “Telecommunications of the Future” (PE00000001 - program “RESTART”, Structural Project 6GWINET, Cascade Call Project SPARKS).
G. Di Gennaro was supported by the Italian Ministry for University and Research (MUR) - PON Ricerca e Innovazione 2014–2020 (D.M. 1062/2021). 
S. Buzzi was also supported by Horizon Europe project CENTRIC (Grant No. 101096379). 
F.A.N. Palmieri was also supported by POR CAMPANIA FESR 2014/2020, A-MOBILITY: Technologies for Autonomous Vehicles.}
\thanks{G. Di Gennaro, G. Romano and F.A.N. Palmieri are with Dipartimento di Ingegneria, Università degli Studi della Campania ``Luigi~Vanvitelli'', Aversa (CE), 81031, Italy (e-mails:~\{giovanni.digennaro; gianmarco.romano; francesco.palmieri\}@unicampania.it).}
\thanks{A. Buonanno is with the Department of Energy Technologies and Renewable Sources, ENEA, Portici (NA), 80055, Italy (e-mail: amedeo.buonanno@enea.it).}
\thanks{S. Buzzi is with University of Cassino and Southern Lazio, I-03043 Cassino, Italy, with Consorzio Nazionale Interuniversitario per le Telecomunicazioni, I-43124 Parma, Italy, and with Politecnico di Milano, I-20133 Milano, Italy (e-mail: buzzi@unicas.it).} 
}

\markboth{IEEE TRANSACTIONS ON COMMUNICATIONS,~Vol.~XX, No.~X, ...}%
{Di Gennaro \MakeLowercase{\textit{(et al.)}}: A General Framework for Scalable UE-AP Association in User-Centric CF-mMIMO based on LSTM NNs}

\maketitle

\begin{abstract}
This study addresses the challenge of access point (AP) and user equipment (UE) association in cell-free massive MIMO networks. 
It introduces a deep learning algorithm leveraging Bidirectional Long Short-Term Memory cells and a hybrid probabilistic methodology for weight updating.
This approach enhances scalability by adapting to variations in the number of UEs without requiring retraining. 
Additionally, the study presents a training methodology that improves scalability not only with respect to the number of UEs but also to the number of APs. 
Furthermore, a variant of the proposed AP-UE algorithm ensures robustness against pilot contamination effects, a critical issue arising from pilot reuse in channel estimation. 
Extensive numerical results validate the effectiveness and adaptability of the proposed methods, demonstrating their superiority over widely used heuristic alternatives.                 
\end{abstract}


\section{Introduction}
\IEEEPARstart{I}{n} recent years, cell-free massive MIMO (CF-mMIMO) \cite{Ngo2015,Ngo2017,Demir2021} has emerged as a promising network deployment paradigm for future sixth-generation (6G) wireless systems. 
Unlike traditional cellular architectures, CF-mMIMO leverages a dense distribution of low-complexity access points (APs) across the coverage area, all interconnected via high-capacity fronthaul links to one or more central processing units (CPUs).
These APs operate cooperatively to serve all user equipments (UEs) in a non-cellular fashion, allowing each UE to simultaneously benefit from multiple APs. 
In particular, within the scalable user-centric formulation of CF-mMIMO \cite{buzzi2017cell,Interdonato2019,Bjornson2020,Buzzi2020, Interdonato2024}, each UE is dynamically associated with a small subset of nearby APs, effectively forming a personalized UE-specific virtual cell.
One of the key advantages of CF-mMIMO over conventional massive MIMO is its ability to provide more uniform performance across all users, effectively mitigating the cell-edge issues that often degrade service in cellular networks. 
By eliminating fixed cell boundaries, CF-mMIMO ensures a consistent quality of service (QoS) regardless of a UE's location \cite{interdonato2019ubiquitous}.
Additionally, CF-mMIMO enhances link reliability and efficiency. 
With APs positioned closer to UEs than in centralized massive MIMO systems, the network benefits from reduced latency, lower path loss, and decreased power consumption. 
In low-load scenarios, deactivating some APs further improves energy efficiency. 
Moreover, connecting each UE to multiple APs introduces macro-diversity, significantly increasing resilience against signal blockages and fading.

\begin{figure}[!t]
    \centering
    \includegraphics[width=0.8\linewidth]{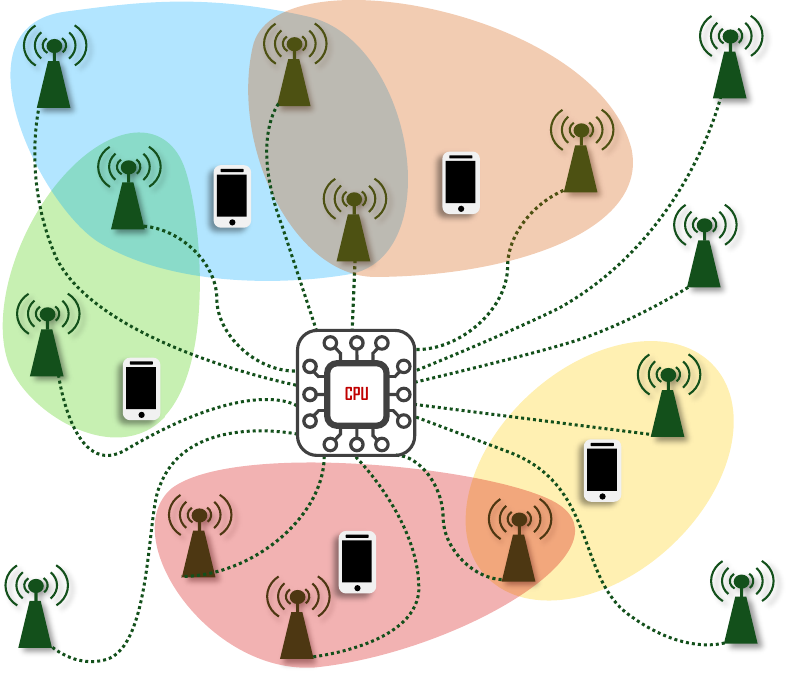}
    \caption{A user-centric CF-mMIMO paradigm illustrating multiple low-complexity APs serving a set of UEs within the same time-frequency slot, all interconnected through fronthaul links to a central processing unit (CPU).}
    \label{fig:cellfree}
\end{figure}

Unlike the theoretical CF-mMIMO model, where every AP can potentially serve every UE, recent research has increasingly shifted towards a user-centric approach \cite{Ammar2022, Buzzi2020}. 
In this approach (Figure~\ref{fig:cellfree}), each UE is served by a carefully selected subset of APs, which helps reduce computational complexity and enhances overall system performance. 
However, this paradigm introduces a new layer of complexity regarding the optimal association between APs and UEs.
This association problem is inherently combinatorial and quickly becomes computationally intractable, even in small networks, necessitating scalable and efficient alternatives.
Consequently, these subsets are typically determined using heuristic methods based on factors such as channel quality, distance, and other relevant parameters \cite{Ammar2019, Bjornson2020, Buzzi2020}. 
While these heuristic solutions provide useful baseline performance, they often lack the adaptability required for specific contexts, leading to inefficiencies and suboptimal results.
This paper addresses this challenge by proposing a learning-based approach that generalizes the AP-UE association problem.
Our method is dynamically adaptable to varying network sizes and establishes a comprehensive AP-UE association framework capable of optimizing multiple network objectives, thereby enhancing efficiency and performance across diverse scenarios.

The current paper is an extended version of the earlier study \cite{GDG_SPAWC2024}, which presented initial results on the design of scalable AP-UE association rules based on deep learning. 

%
%
%
%
%
%
\subsection{State of the art}
Numerous studies have been conducted on AP selection and clustering in CF-mMIMO networks, highlighting the significance of this problem in optimizing network performance. 
Among the most relevant approaches, classical clustering methods such as K-means and Gaussian Mixture Models (GMM) have been widely explored. 
For example, K-means++ clustering has been applied in \cite{Biswas2021} to develop a cluster-based AP selection algorithm, mitigating pilot contamination and balancing the computational workload in CF-mMIMO systems. 
Similarly, user mobility-aware clustering based on K-means++ has been investigated in \cite{liu2024user}, where adaptive AP selection strategies help reduce inter-AP switching and improve network throughput in high-mobility scenarios. 
Additionally, GMM-based clustering, as introduced in \cite{biswas2024optimal}, aims to optimize the trade-off between cluster size and data rates. 
However, despite their relevance, these approaches require defining the number of clusters in advance and/or rely on the assumption that each cluster must always serve a fixed number of users. 
This assumption is flawed because (unless the problem is simplified by forcing each UE to connect to a single AP) it results in APs being compelled to serve distant users, leading to inefficiencies due to the initial heuristic choices.

Other methodologies have attempted to tackle AP selection by leveraging game-theoretic frameworks and information rate-based clustering. 
In \cite{wei2025game}, a game-theoretic approach is introduced, recognizing the impact of AP selection on overall service quality. 
These models, while insightful, often leave the decision in the hands of the users, assuming non-opportunistic behavior, which may not hold in realistic network scenarios. 
A different perspective is taken in \cite{rachuri2024novel}, where joint user association and power control are optimized to enhance spectral efficiency while ensuring fairness. 
Similarly, clustering methods based on information rates, such as those in \cite{mashdour2024clustering}, prioritize fairness and efficiency.
Although very interesting, these approaches do not directly tackle the clustering problem itself but rather focus on meeting specific performance requirements during clustering, excluding users who do not satisfy them and deferring their connection to future time slots.
In the same vein, additional studies have concentrated on developing clustering solutions specifically designed to address the unique challenges posed by 5G. 
For instance, \cite{wang2024user} examines how ergodic rate-based clustering can improve fairness and reliability in networks that must meet ultra-reliable and low-latency communication (URLLC) requirements. 
These studies highlight the need for clustering strategies that extend beyond classical performance metrics to address the critical reliability and latency constraints of next-generation wireless systems.

More recently, deep reinforcement learning (DRL)-based clustering techniques have emerged as an alternative. 
For instance, in \cite{tsukamoto2023user}, DRL has been used to formulate user-centric AP clustering as a Markov Decision Process, allowing APs to independently learn optimal user assignments. 
Similarly, \cite{Mendoza2023} introduces a DRL framework to dynamically adjust clustering, either to satisfy user-specific demands or to maximize spectral efficiency. 
Multi-agent reinforcement learning (MARL) and federated MARL (MAFRL) have also been explored in \cite{Banerjee2023} to enable autonomous AP learning and improve adaptability in mobile environments. 
However, these methods typically rely on centralized subnetworks and, more importantly, lack scalability with respect to the number of users and/or APs. 
As a result, when the network expands, these approaches require retraining, making them impractical for large-scale and dynamically evolving CF-mMIMO deployments.

\subsection{Contributions}
This paper presents a \emph{Deep Learning} (DL) approach for distributed AP-UE association in a CF-mMIMO system.
Through the strategic use of \emph{Bidirectional Long Short-Term Memory} (BiLSTM) networks, this paper proposes an innovative association algorithm that, \emph{differently from previous research}, ensure scalability with respect to the number of UEs and APs, while also demonstrating robustness against pilot contamination effects.
More specifically, the paper contribution can be summarized as follows.
\begin{itemize}
    \item[-] The proposed approach employs DL to integrate a BiLSTM network that generalizes the clustering problem, enabling the dynamic reconfiguration of AP-UE connections based on channel gain inputs. 
    By determining learning through a \emph{probabilistic framework}, this approach significantly enhances the ability to predict and optimize connectivity with high adaptability, ultimately improving overall performance in fluctuating network conditions.
    \item[-] Leveraging this architecture, the method adopts a master-centric strategy that centralizes decision-making within a single AP. 
    This approach provides the AP with a comprehensive view of the network's state, enabling efficient monitoring and ensuring compliance with operational constraints. 
    Notably, it is precisely this master-centric design, when integrated with the RNN structure, that enables the system to scale effectively with the number of UEs. 
    By concentrating decision-making at a single AP, the network can accommodate an increasing number of UEs without the need for retraining, ensuring that the system remains responsive and efficient in dynamic environments.  
    \item[-] To ensure broad applicability across various scenarios, the method is evaluated with three distinct loss functions, each targeting critical objectives within cell-free environments. 
    This multi-objective evaluation underscores the method's generality and effectiveness in addressing diverse performance metrics, making it adaptable to different operational goals and enhancing its overall utility.
    \item[-] In addition to these features, it is demonstrated that the proposed training methodology can be made scalable also with respect to the number of APs while becoming fully parallelizable in a distributed manner.
    This flexibility enables the system to efficiently adapt to varying network topologies, thereby enhancing its suitability for large-scale deployments.
    \item[-] Finally, a comprehensive numerical performance analysis is conducted to assess the robustness and adaptability of the proposed approach across diverse scenarios, verifying that it consistently outperforms conventional methods on each specific objective of all loss functions.
\end{itemize}

The paper is organized as follows.
In the next section, a brief overview of preliminary concepts related to LSTM networks is provided, establishing the foundation for understanding their role in this study.
Section \ref{sec:cell-free} then introduces the user-centric CF-mMIMO model implemented for the communication channel, detailing its architecture and operational principles.
In Section \ref{sec:problem}, the problem is formally defined, with an explicit formulation of the constraints and a description of the three loss functions used in the analysis.
Section \ref{sec:DL} presents the proposed BiLSTM-based deep learning approach, highlighting its architecture and training methodology.
Finally, Section \ref{sec:simulations} discusses the simulation results, demonstrating the performance and effectiveness of the proposed model.

\begin{figure}[!b]
    \centering
    \includegraphics[width=\linewidth]{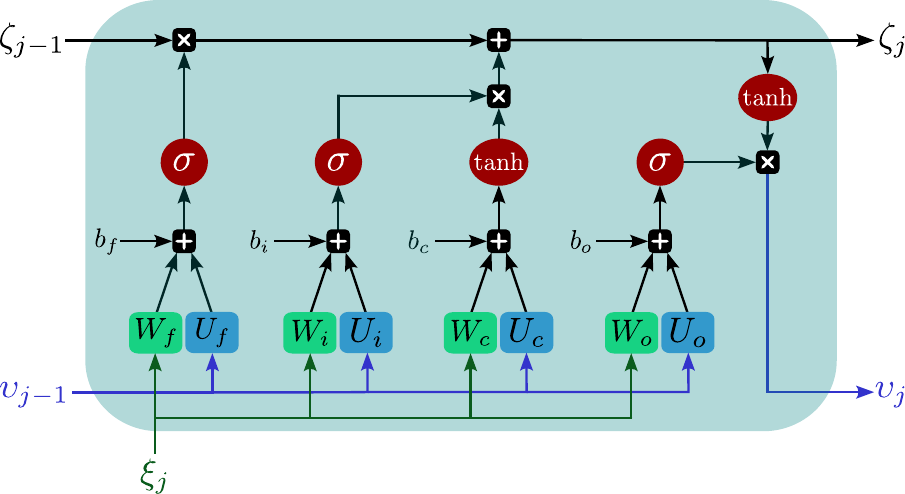}
    \caption{Conceptual scheme of an LSTM cell.}
    \label{fig:LSTM_cell}
\end{figure}

\section{Preliminaries on LSTM networks}
\label{sec:lstm}
\emph{Long Short-Term Memory} (LSTM) networks are a specialized type of \emph{Recurrent Neural Network} (RNN) designed to address the challenge of maintaining long-term dependencies in sequential data. 
Unlike standard RNNs, which struggle with vanishing gradient issues over long sequences, LSTMs incorporate memory cells and gate mechanisms to selectively retain, forget, or update information over time.
This ability to effectively manage the flow of information enables them to capture complex temporal patterns, making them particularly suited for tasks such as natural language processing \cite{DiGennaro2022_ijcnn}, time-series prediction, and audio analysis \cite{Scarpiniti2020}. 
Their structured approach to handling sequential dependencies has established LSTMs as a fundamental tool in various machine learning applications requiring context awareness and long-range information retention.

The architecture of an LSTM cell is depicted in Figure~\ref{fig:LSTM_cell}. 
At each time step $j$, the cell receives an \emph{input vector} $\pmb{\xi}_j$ and updates both its \emph{hidden state} 
$\pmb{\upsilon}_{j}$ and \emph{cell state} $\pmb{\zeta}_j$.
These updates are governed by the following equations:
\begin{align*}
    \mathbf{f}_j &= \varsigma(\mathbf{W}_f\,\pmb{\xi}_j + \mathbf{U}_f\,\pmb{\upsilon}_{j-1} + \mathbf{b}_f) \\
    \mathbf{i}_j &= \varsigma(\mathbf{W}_i\,\pmb{\xi}_j + \mathbf{U}_i\,\pmb{\upsilon}_{j-1} + \mathbf{b}_i) \\
    \mathbf{o}_j &= \varsigma(\mathbf{W}_o\,\pmb{\xi}_j + \mathbf{U}_o\,\pmb{\upsilon}_{j-1} + \mathbf{b}_o) \\
    \mathbf{c}_j &= \tanh(\mathbf{W}_c\,\pmb{\xi}_j + \mathbf{U}_c\,\pmb{\upsilon}_{j-1} + \mathbf{b}_c)
\end{align*}
where $\varsigma(\cdot)$ is the sigmoid (standard logistic) activation function. 
The learnable parameters include the weight matrices $\mathbf{W}_* \in \mathbb{R}^{q \times d}$ and $\mathbf{U}_* \in \mathbb{R}^{q \times q}$, along with the bias vectors $\mathbf{b}_* \in \mathbb{R}^q$.
It is important to note that, during the simulation phase, the only parameter that needs to be specified is the dimension $q$ of the LSTM’s hidden state, since the input dimension $d = \dim(\pmb{\xi})$ is inherently determined by the problem definition.
The hidden state vector $\pmb{\upsilon}_j$ is ultimately derived from the cell state $\pmb{\zeta}_j$ as follows:
\begin{align*}
    \pmb{\zeta}_j &= \mathbf{f}_j \odot \pmb{\zeta}_{j-1} + \mathbf{i}_j \odot \mathbf{c}_j \\
    \pmb{\upsilon}_j &= \mathbf{o}_j \odot \tanh(\pmb{\zeta}_j)
\end{align*}
where the operator $\odot$ denotes the Hadamard product, indicating element-wise multiplication. 
The vectors $\pmb{\zeta}_{j-1}$ and $\pmb{\upsilon}_{j-1}$ together provide the contextual information for the next cell $j$, subsequently affecting its output. 
In this work, as is typical in many LSTM applications, the initial values $\pmb{\zeta}_0$ and $\pmb{\upsilon}_0$ for the first AP are set to zero vectors. 
This initialization does not introduce any prior into the model, allowing the LSTM to construct its representations based solely on the sequence of input data, while still being dependent on the previous state.

\begin{figure}[!t]
    \centering
    \includegraphics[width=\linewidth]{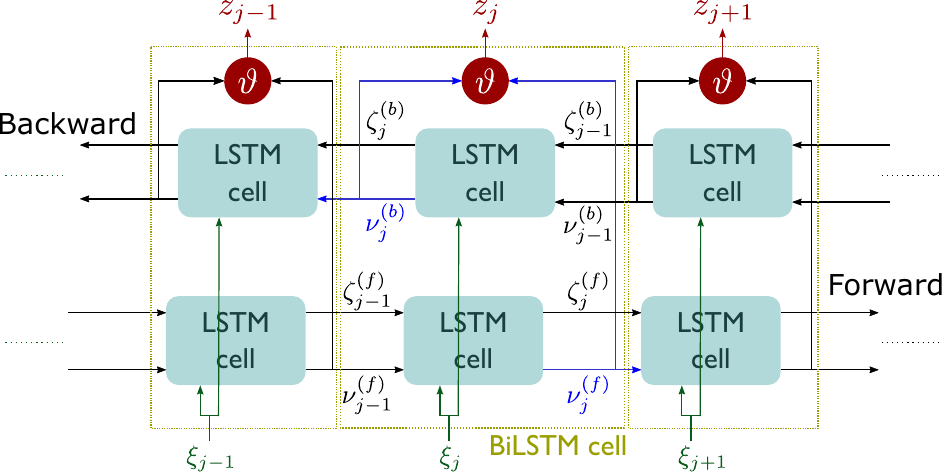}
    \caption{Conceptual scheme of a BiLSTM network.}
    \label{fig:BiLSTM_scheme}
\end{figure}

As previously discussed, the AP-UE association algorithm presented in this paper is fundamentally based on the bidirectional variant of LSTM, known as BiLSTM. 
Unlike traditional LSTM architectures, the BiLSTM processes sequences in both forward and backward directions, allowing the model to effectively capture richer contextual information (see Figure~\ref{fig:BiLSTM_scheme}).
This bidirectional processing is valuable for tasks relying on context from both ends, closely resembling classical probabilistic approaches \cite{Palmieri2022, Buonanno2021}.
The BiLSTM network integrates both Forward and Backward LSTMs to handle sequential data, enabling it to grasp contextual nuances and improve its ability to model intricate dependencies in the data.
In the Forward LSTM, each cell processes the inputs in the order defined by the sequence. 
At each $j$-th step, it maintains an internal state ({\footnotesize $\pmb{\zeta}^{(f)}_j$}) and generates a hidden state ({\footnotesize $\pmb{\upsilon}^{(f)}_j$}). 
Conversely, in the Backward LSTM, the cell operates in reverse order compared to its Forward LSTM counterpart. 
At each $j$-th step, it also maintains its own internal state ({\footnotesize $\pmb{\zeta}^{(b)}_j$}) and produces a hidden state ({\footnotesize $\pmb{\upsilon}^{(b)}_j$}) for each element of the input.
This reverse processing enables the network to utilize context from subsequent elements, enriching the information captured from preceding elements through the Forward LSTM. 
A key aspect of this architecture is the merging of hidden state vectors from both LSTMs for each position in the sequence. Specifically, the BiLSTM cell's output at the $j$-th position is given by:
\begin{equation*}
    \pmb{z}_j = \vartheta\left(\pmb{\upsilon}^{(f)}_j, \pmb{\upsilon}^{(b)}_j\right)
\end{equation*}
where $\vartheta(\cdot)$ represents the aggregation function used to combine hidden states from Forward and Backward LSTM cells.

\section{User-Centric CF-mMIMO model and transceiver processing}
\label{sec:cell-free}
Our study examines a downlink scenario involving $K$ single-antenna UEs and $L$ APs, all distributed over the same geographic area. 
Each AP is equipped with $M$ antennas, and the system operates in an \emph{ultra-dense} regime, characterized by a number of APs significantly exceeding the number of UEs ($L \gg K$) \cite{Ngo2015}. 
In this setting, the cumulative number of antennas across all APs is considerably larger than the number of UEs ($L \times M \gg K$), enhancing spatial multiplexing capabilities.
However, since each AP is equipped with a limited number of antennas, it may end up serving more UEs than it can effectively handle.
As a result, local capabilities of individual APs are insufficient to fully exploit the massive MIMO benefits, making coordination among APs essential for effectively managing inter-user interference.

Communication between APs and UEs relies on the \emph{Time-Division Duplex} (TDD) mode, where uplink and downlink transmissions share the same frequency band but occur in alternating time slots.
Assuming that the channel coherence time is sufficiently long compared to the symbol transmission rate, TDD facilitates channel estimation through reciprocity, as the channel remains approximately constant within each coherence block.
In particular, each coherence block consists of $\tau_c$ symbols, allocated as follows: $\tau_p$ symbols for uplink pilots, $\tau_u$ symbols for uplink data transmission, and $\tau_d$ symbols for downlink data transmission, such that $\tau_c = \tau_p + \tau_u + \tau_d$.

To simulate non-line-of-sight (NLoS) propagation channels, we represent the channel between AP $\ell \in \mathcal{L}=\{1,\dots,L\}$ and UE $k \in \mathcal{K}=\{1,\dots,K\}$ by: 
\begin{equation}
    \mathbf{h}_{k\ell} = \sqrt{\beta_{k\ell}} \tilde{\mathbf{h}}_{k\ell} \in \mathbb{C}^M
\end{equation}
where $\tilde{\mathbf{h}}_{k\ell} \sim \mathcal{N}_\mathbb{C}(\mathbf{0}_M, \mathbb{I}_M)$ models small-scale fading, and $\beta_{k\ell}$ reflects the large-scale fading variance \cite{Demir2021}.

For network connectivity, it is assumed that each AP periodically broadcasts synchronization signals, which UEs use to estimate the slowly varying channel gains.
Based on these estimates, the $k$-th UE identifies its \emph{master AP} by selecting the AP with the highest channel gain, as expressed by:
\begin{equation}
    m_k = \argmax_{\ell \in \mathcal{L}} \beta_{k\ell}
    \label{eq:master}
\end{equation}
Once the master AP is identified, UE $k$ can directly communicate with it (e.g., via a conventional random access method) to obtain a pilot assignment and initiate data transfer.

\paragraph{Master selection} 
The network is designed to utilize $\tau_p$ mutually orthogonal pilot sequences $\phi_1, \dots, \phi_{\tau_p}$, each satisfying $\|\phi_t\|^2 = \tau_p$ for all $t \in \mathcal{T}=\{1,\dots,\tau_p\}$. 
For simplicity, we assume that the master AP is responsible for locally selecting pilots for its associated UEs. 
As outlined in \cite{Bjornson2020}, this approach reduces pilot interference within the AP by assigning the $k$-th UE the pilot $\phi_{t_k}$, with the index $t_k$ selected according to:
\begin{equation}
    t_k = \argmin_{t \in \mathcal{T}} \sum_{i \in \mathcal{P}_t} \beta_{i m_k}
\end{equation}
where $\mathcal{P}_t \subset \mathcal{K}$ represents the set of UEs sharing the $t$-th pilot.
Once the pilot sequence has been assigned, each master AP informs the network via the CPU, ensuring all APs are notified of the newly connected UE.
During this process, regardless of whether the system operates in a centralized or distributed manner, the available information is evaluated to determine whether each AP should establish a connection with the specific UE.
Ultimately, the subset $\mathcal{L}_k \subset \mathcal{L}$ of APs responsible for managing communication with UE $k$ will be created.

\paragraph{Channel estimation}
Since the TDD protocol is synchronized across all APs, the transmission of pilots from UEs connected to the network will reliably occur within $\tau_p$ samples of the subsequent coherence block. 
For coherent transmission to occur, each AP $\ell \in \mathcal{L}_k$ must estimate the channel vector $\mathbf{h}_{k\ell}$ by evaluating the received signal $Y_\ell^\mathrm{pilot} \in \mathbb{C}^{M \times \tau_p}$ during this phase.
This signal can be expressed as follows:
\begin{equation}
	Y_\ell^\mathrm{pilot} = \sum_{i=1}^K \sqrt{\eta_i} \mathbf{h}_{i\ell} \phi_{t_i}^\top + \mathbf{N}_\ell \qquad
\end{equation}
where $\mathbf{N}_\ell \in \mathbb{C}^{M \times \tau_p}$ represents the noise at the receiver, with i.i.d. elements distributed according to $\mathcal{N}_\mathbb{C}(0, \sigma_{\mathrm{ul}}^2)$, and $\eta_i \ge 0$ denotes the uplink power of the $i$-th UE.
The channel estimate is derived by first eliminating interference from the orthogonal pilots, achieved by multiplying the received signal by the normalized conjugate of the corresponding pilot $\phi_{t_k}$, yielding:
\begin{align*}
    \mathbf{y}_{t_k \ell}^\mathrm{pilot} &= \frac{1}{\sqrt{\tau_p}} Y_\ell^\mathrm{pilot} \phi^{*}_{t_k} \\
    &= \underbrace{\vphantom{\sum_{i \in \mathcal{P}_{t_k}}}\sqrt{\tau_p \eta_k} \mathbf{h}_{k\ell}}_\text{Desired} + \underbrace{\sqrt{\tau_p}\!\!\!\sum_{i \in \mathcal{P}_{t_k} \setminus \{k\}}\!\!\!\sqrt{\eta_i} \mathbf{h}_{i\ell}}_\text{Interference} + \underbrace{\vphantom{\sum_{i \in \mathcal{P}_{t_k}}} \mathbf{n}_{t_k \ell}}_\text{Noise}
\end{align*}
where the elements of the noise vector $\mathbf{n}_{t_k \ell} \sim \mathcal{N}_\mathbb{C}(\mathbf{0}_M, \sigma_{\mathrm{ul}}^2 \mathbf {I}_M)$ are still i.i.d. since $\phi^{*}_{t_k} / \sqrt{\tau_p}$ is a unit norm vector.

The MMSE channel estimate can now be derived via:
\begin{equation}
    \mathbf{\hat{h}}_{k\ell} = \frac{\gamma_{k\ell}}{\sqrt{\tau_p \eta_k} \beta_{k\ell}} \mathbf{y}_{t_k \ell}^\mathrm{pilot}
\end{equation}
where we have defined 
\begin{equation*}
    \gamma_{k\ell} = \frac{\tau_p \eta_k \beta_{k\ell}^2}{\displaystyle\tau_p\!\!\sum_{i \in \mathcal{P}_{t_k}}\!\!\eta_i \beta_{i\ell} + \sigma_{\mathrm{ul}}^2}.
\end{equation*}
Note that, given the independence of the channels, channel estimates can be calculated individually at each AP without compromising optimality. 

\paragraph{Downlink operation}
Assuming channel reciprocity, the downlink signal received by UE $k$ can be expressed as follows:
\begin{equation}
    y_k^\mathrm{dl} = \underbrace{\vphantom{\sum_{\setminus}}s_k \sum_{\ell \in \mathcal{L}_k} \mathbf{h}_{k\ell}^\mathrm{H} \pmb{\omega}_{k\ell}}_\text{Desired} + 
    \!\!\underbrace{\sum_{i \in \mathcal{K} \setminus \{k\}}\!\!\!\!s_i \sum_{\ell \in \mathcal{L}_i} \mathbf{h}_{k\ell}^\mathrm{H} \pmb{\omega}_{i\ell}}_\text{Interference} +
    \underbrace{\vphantom{\sum_{\setminus}}n_k}_\text{Noise}
\end{equation}
where $s_k \in \mathbb{C}$ is the unity power downlink data signal for the $k$-th UE ($\mathbb{E}\{|s_i|^2\} = 1$), $n_k \sim \mathcal{N}_\mathbb{C}(0, \sigma_\mathrm{dl}^2)$ represents the noise at the receiver, and $\pmb{\omega}_{k\ell} \in \mathbb{C}^M$ is the effective precoding vector.
Since the selection of the precoding vector involves two distinct subtasks (directivity and power allocation), it is typically expressed for each AP $\ell$ serving UE $k$ as follows:
\begin{equation}
    \pmb{\omega}_{k\ell} = \sqrt{\rho_{k\ell}} \frac{\pmb{\bar{\omega}}_{k\ell}}{\sqrt{\mathbb{E}\{\|\pmb{\bar{\omega}}_{k\ell}\|^2\}}}
\end{equation}
where $\rho_{k\ell} \ge 0$ is the transmit power assigned by AP $\ell$ to UE $k$, and $\pmb{\bar{\omega}}_{k\ell}$ is a scaled vector indicating transmission directivity.

In a scalable network with distributed beamforming and channel estimation, the downlink \emph{Spectral Efficiency} (SE) for UE $k$ is expressed as:
\begin{equation}
	\mathrm{SE}_k = \frac{\tau_d}{\tau_c} \log_2 \left(1 + \mathrm{SINR}_k\right) \qquad \unit[per-mode = repeated-symbol]{\bit \per \s \per \Hz}
\end{equation}
where the pre-log factor $\tau_d / \tau_c$ denotes the portion of each coherence block used for data transmission, and the effective \emph{Signal-to-Interference-plus-Noise Ratio} (SINR) is given by
\begin{equation*}
	\mathrm{SINR}_k = \frac{\displaystyle\left|\sum_{\ell \in \mathcal{L}_k}\!\! \mathbb{E}\{\mathbf{h}_{k\ell}^\mathsf{H} \pmb{\omega}_{k\ell}\} \right|^2}{\displaystyle\sum_{i \in \mathcal{K}} \mathbb{E}\!\left\{\!\left|\sum_{\ell \in \mathcal{L}_i}\!\! \mathbf{h}_{k\ell}^\mathsf{H} \pmb{\omega}_{i\ell} \right|^2\!\right\} - \left| \sum_{\ell \in \mathcal{L}_k}\!\!\mathbb{E}\{\mathbf{h}_{k\ell}^\mathsf{H} \pmb{\omega}_{k\ell}\} \right|^2\! + \sigma_\mathrm{dl}^2}.
\end{equation*}
In this context, a possible approach for determining $\mathbf{w}_{k\ell}$ is to employ \emph{Maximum Ratio} (MR) precoding, which is achieved for each AP $\ell \in \mathcal{L}_k$ by simply setting $\pmb{\bar{\omega}}_{k\ell} = \mathbf{\hat{h}}_{k\ell}$.
This choice also allows for closed-form expectations in $\mathrm{SINR}_k$, yielding:
\begin{equation*}
    \mathrm{SINR}_k = \frac{\displaystyle N \left(\sum_{\ell \in \mathcal{L}_k} \sqrt{\rho_{k\ell} \gamma_{k\ell}} \right)^{\!\!2}}{\displaystyle \sum_{i \in \mathcal{K}} \sum_{\ell \in \mathcal{L}_i} \rho_{i\ell}\beta_{k\ell} + N\!\!\!\!\!\sum_{i \in \mathcal{P}_k \setminus\{k\}}\!\!\!\left(\sum_{\ell \in \mathcal{L}_i} \sqrt{\rho_{i\ell} \gamma_{k\ell}} \right)^{\!\!2} + \sigma_\mathrm{dl}^2}\, .
\end{equation*}
For the sake of brevity, this paper just considers MR precoding, and the analysis of the impact on the system performance of other precoders is not taken into account. 

\section{Problem formulation}
\label{sec:problem}
As previously stated, the main objective of this work is to develop scalable AP-UE association rules that optimize a specific performance metric.
Mathematically, we aim to address the following constrained optimization problem:
\begin{maxi!}
    {\scriptstyle\mathcal{L}_1, \dots, \mathcal{L}_K}{\psi( \mathcal{L}_1, \ldots, \mathcal{L}_K) \label{eq:objectiveProblem}}{\label{eq:Problem}}{} \addConstraint{\mathcal{L}_k}{\ne\emptyset \quad}{\forall k \in \mathcal{K} \label{eq:C1Problem}}
    \addConstraint{\sum_{i \in \mathcal{K}_\ell} \rho_{i\ell}}{\leq \rho_\mathrm{max} \qquad}{\forall \ell \in \mathcal{L} \label{eq:C2Problem}}
\end{maxi!}
where $\mathcal{K}_\ell \subset \mathcal{K}$ denotes the subset of UEs managed by AP $\ell$, and $\psi(\cdot)$ is a general objective function to be maximized, allowing the problem to be tailored to specific goals.
Since our focus is on optimizing this function with respect to the AP-UE association, we have exclusively underscored the dependence on the sets $\mathcal{L}_1, \dots, \mathcal{L}_K$.
Additionally, constraint \eqref{eq:C1Problem} ensures that each UE is connected to at least one AP, to guarantee that no UE remains disconnected, while constraint \eqref{eq:C2Problem} guarantees that the total transmitted power at each AP does not exceed the maximum transmission power $\rho_\mathrm{max}$, which we assume to be the same for all APs.

As detailed in the following, several choices are possible for the objective function $\psi(\cdot)$, which demonstrates the versatility of the proposed approach in catering to diverse requirements and operational contexts.

\paragraph{Cumulative Weighted Spectral Efficiency}
The simplest method is to aim at enhancing overall spectral efficiency, which can be achieved by defining:
\begin{equation}
    \psi_\mathrm{SUM}(\cdot) = \sum_{k \in \mathcal{K}} \alpha_k \mathrm{SE}_k \; ,
    \label{eq:psi_sum}
\end{equation}
where $\alpha_k\geq 0$ is a scalar that assigns a weight to the relevance of user $k$ within the system.  
Setting $\alpha_k=1$ for all $k\in \mathcal{K}$, the expression in \eqref{eq:psi_sum} simplifies to the sum of the spectral efficiencies of all users, representing the total volume of downlink data transmitted by the network across the available spectrum.

\paragraph{Spectral Efficiency vs. Number of Connections}
In CF-mMIMO systems, handling multiple AP-UE connections involves intricate tasks such as power control, beamforming, and signal processing. 
Each connection demands backhaul resources and coordination; thus, minimizing active connections per user can lead to improved efficiency and saving of useful resources. 
To address these needs without significantly compromising system performance, it is essential to strike a balance between improving spectral efficiency and minimizing the number of connections.
This multi-objective challenge can be tackled by formulating the following objective function:
\begin{equation}
    \psi_\mathrm{BALANCE}(\cdot) = \sum_{k \in \mathcal{K}} (\mathrm{SE}_k - \lambda |\mathcal{L}_k|)
    \label{eq:psi_balance}
\end{equation}
where $|\mathcal{L}_k|$ denotes the size of the set $\mathcal{L}_k$, and $\lambda$ is a fixed (i.e., non-learned) positive weighting parameter. 
Clearly, a higher $\lambda$ prioritizes minimizing the number of connections, leading to a simpler system but possibly sacrificing spectral efficiency and overall throughput.
Conversely, a lower $\lambda$ places greater emphasis on maximizing SE, enabling more APs to serve each user, albeit with increased system complexity.

\paragraph{Minimum Spectral Efficiency}
Unlike traditional cellular networks, where users near the base station experience better signal quality than those at the cell's edge, cell-free networks aim to ensure fairness by using multiple distributed APs. 
However, even in this scenario, some users may still encounter suboptimal performance, particularly those located in high-interference areas.
In situations where it is crucial to ensure that these users also receive an adequate quality of service, the following objective function can be utilized:
\begin{equation}
	\psi_\mathrm{MIN}(\cdot) = \min_{k \in \mathcal{K}} \mathrm{SE}_k \; .
    \label{eq:psi_min}
\end{equation}
Maximizing the minimum spectral efficiency seeks thus to mitigate the poor performance experienced by a subset of disadvantaged users. 
However, this approach inherently involves trade-offs that must be considered from the outset. 
Enhancing the performance of the weakest users necessitates some reduction in the performance of the strongest ones, thereby limiting improvements in cumulative spectral efficiency and the total number of connections achieved.

It is essential to emphasize that the three functions presented above are merely illustrative examples used to generate numerical results.
Our approach is inherently general, accommodating a wide range of alternative options. 
For instance, the function $\psi_\mathrm{MIN}(\cdot)$ can be modified by incorporating a penalty term, akin to the definition in \eqref{eq:psi_balance}, to achieve a trade-off between the minimum SE among users and the number of active connections. 
Additionally, global energy efficiency \cite{buzzi2016survey} (defined as the total SE across users divided by the overall network power consumption) can be employed to optimize the transmission of bits per unit of energy in the system. 
However, for the sake of brevity, this paper does not delve into these and other potential objective functions.

\section{Scalable AP-UE association based on DL}
\label{sec:DL}
To introduce our scalable DL approach for addressing the AP-UE association problem defined in \eqref{eq:Problem}, we find it more appropriate to separate the various concepts.
Specifically, we will first define how the solution implements BiLSTM cells and describe the methodology that allows it to scale with the number of UEs. 
Subsequently, we will extend this solution to demonstrate how it can be made fully parallelizable in a distributed manner, also scaling with the number of APs. 
More precisely, the final solution we present is based on the assumption that as the system scales up with a fixed ratio between the number of UEs and APs, the number of UEs managed by each AP remains limited. 
This guarantees that the complexity for each AP remains finite, even as the network size expands indefinitely.

\subsection{Scalable Solution for UEs}
In tackling the combinatorial problem \eqref{eq:Problem}, our goal is to develop a system that not only adheres to the necessary constraints but also adapts to changes in the number of UEs \emph{without requiring retraining or modifications to the network architecture}. 
To enable this adaptability, we introduce a \emph{master-centric} strategy in which only the master AP for each UE determines which APs should connect to that UE.
In particular, each master AP will assume full responsibility for assembling the complete cluster for the UEs under its management.
Consequently, this approach also leads to an equitable distribution of computational workload among APs, assuming a uniform user distribution across the coverage area.

In this version, the network architecture (detailed later) comprises all master APs working together to form a unified BiLSTM network, with each master AP generating a BiLSTM cell for each UE under its management.
The outputs of these cells connect to a shared neural structure across all APs, producing an $L$-dimensional vector that, when binarized, indicates the APs to which a specific UE should connect.
Consequently, the total number of BiLSTM cells in the network corresponds to the total number of UEs in the system, while each master AP only implements the cells for the UEs it manages. 
If an AP accepts to become a master of a new UE, it must simply adds another BiLSTM cell with the same architecture as its existing ones.
Note that because the terminal neural structure is shared among all APs and due to the intrinsic properties of BiLSTM cells, the weights associated with each UE’s segment of the network are inherently identical, allowing the network to replicate its structure indefinitely.
Consequently, since each portion of the network operates exclusively for its assigned UE, the overall network performance remains unaffected by the number of UEs, thus ensuring scalability in this regard.

\begin{figure}[!b]
    \centering
    \includegraphics[width=\linewidth]{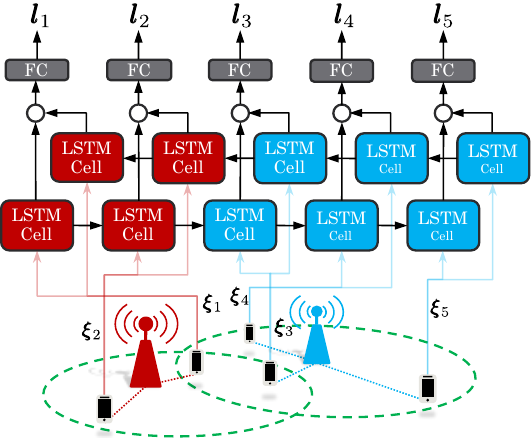}
    \caption{Deep learning framework using UE ordering by master APs.}
    \label{fig:DL}
\end{figure}

Before delving into the specifics of the proposed framework, it is essential to emphasize that the first step involves organizing all APs in a specific sequential order, as effective learning in RNNs requires a strict arrangement of input sequences.
This requirement can be easily satisfied since APs are typically installed in fixed locations by the telco operator, making sequential numbering straightforward.
This ordering creates a first hierarchical level (as in \cite{DiGennaro2022, Palmieri2021}), where each UE is assigned to the subchain associated with its respective master AP. 
Within each subchain, UEs are arranged by the AP based on their channel gains, resulting in a second hierarchical level that exhibits quasi-radial patterns relative to the AP’s position.
For clarity, Figure~\ref{fig:DL} illustrates two APs, with the red scheduled first and the blue second, generating subchains of lengths $2$ and $3$, respectively, based on their assigned UEs.
%

\subsubsection{Model Inference}
Recalling that each BiLSTM cell is designed to process information exclusively related to a single UE, and assuming the UEs are ordered sequentially along the chain (such that the $k$-th UE corresponds to the $k$-th cell), the input vector for these cells is defined as:
\begin{equation}
    \pmb{\xi}_k = [\beta_{k1}, \beta_{k2}, \dots, \beta_{kL}, x^{(k)}_{1}, x^{(k)}_{2}]^\top
    \label{eq:inputs}
\end{equation}
where $x^{(k)}_{1}$ and $x^{(k)}_{2}$ are the positions of the UE%
\footnote{Here, we consider 2D coordinates, assuming all UEs are at the same height, but extending this to the 3D case is straightforward.}. 
Note that both the forward and backward LSTM cells within the BiLSTM share the same inputs (Figures \ref{fig:BiLSTM_scheme} and \ref{fig:DL}).
The evaluations conducted by each LSTM cell propagate simultaneously in both backward and forward directions, aggregating at the output. 
To maintain a compact network size (for efficient inference) while meaningfully correlating the two flows with the interference link generated at each step, the sum was selected as the aggregation function of the BiLSTM.
This means that the output at each position is the element-wise sum of the Forward and Backward LSTM hidden states:
\begin{equation}
    \pmb{z}_k = \pmb{\upsilon}^{(f)}_k + \pmb{\upsilon}^{(b)}_k .
\end{equation}

The output vector $\pmb{z}_k$ from each cell group of the BiLSTM is subsequently fed into a multi-layer \emph{Fully Connected} (FC) network, resulting in the final output:
\begin{equation}
    \pmb{l}_k = \varsigma(\varphi(\pmb{z}_k))
\end{equation}
where $\varphi(\cdot)$ generically represents the transformation produced by the FC network, and $\pmb{l}_k$ is the $L$-dimensional output of the FC network.
This network features shared weights across the cells, creating a uniform block that is identical for each AP.
We emphasize that in the final layer of the FC network sigmoid activation functions are applied (element-wise), ensuring that $\pmb{l}_k \in [0,1]^L$. 
This approach aims to calculate the probability that including a specific AP in the cluster will improve performance with respect to the selected metric.
However, during the inference phase, a hard decision must be made on which APs to associate with the UE under consideration.
Specifically, the output cluster $\pmb{\bar{l}}_k \in \{0,1\}^L$ generated for the $k$-th UE during inference is determined by:
\begin{equation}
    \pmb{\bar{l}}_k = \mathcal{H}_\mu(\pmb{l}_k), 
\end{equation}
where $\mathcal{H}_\mu$ denotes the Heaviside function with threshold $\mu$.
In other words, during the inference phase, a threshold $\mu$ determines which link to activate; if the probability exceeds this threshold, the particular AP is included in the cluster for that specific UE.
In this paper, results will be presented by assuming that the threshold is set at $\mu = 0.5$.

It is important to note that the master-centric approach offers a comprehensive local view of the cluster formed for each individual UE, which greatly facilitates any necessary adjustments. 
Since only the master AP is responsible for making decisions regarding a specific UE, it can effectively assess whether any AP has been assigned to that UE. 
This approach also offers the advantage of significantly simplifying the monitoring of compliance with certain requirements, as the master AP can respond appropriately to ensure that these conditions are met. 
For example, regarding the constraint \eqref{eq:C1Problem}, while we address it here in a straightforward manner by assuming that the master AP for each UE is always connected, the proposed approach also allows for the implementation of random or alternative decisions, demonstrating its inherent flexibility.
To clarify the concept, from a fault-tolerant perspective, this approach has the potential to ensure that each user remains connected to at least two APs, a goal that is challenging to achieve with other decentralized, user-centric approaches.

Finally, once the various clusters have been generated, each AP is notified of the specific UEs to which it must connect and can allocate the appropriate transmit power for communicating with each user.
Instead of directly optimizing the transmit power to fulfill constraint \eqref{eq:C2Problem}, this study assigns the power allocated to each UE connected to AP $\ell$ using the method outlined in \cite{Interdonato2019}. 
Specifically, we define:
\begin{equation}
    \rho_{k\ell} = \rho_\mathrm{max} \frac{\sqrt{\beta_{k\ell}}}{\displaystyle\sum_{i \in \mathcal{K}_\ell} \sqrt{\beta_{i\ell}}}\; .
\end{equation}

\subsubsection{Model Training}
The training phase of the model involves a distinct process from the inference phase, as it must also account for the specific objective function being optimized. 
Similar to certain architectures \cite{DiGennaro2021b}, the training process requires the introduction of an auxiliary module tasked with evaluating the specific reward function. 
This secondary element, designed solely to facilitate gradient flow in the primary network, ensures that the weights are adjusted to achieve the intermediate objective (defining the cluster) while remaining aligned with the chosen evaluation metric. 
Once training is complete, this auxiliary unit is discarded, since it is not needed in the final inference phase.

During model training, the need for binary decisions on AP inclusion within each UE’s cluster poses a challenge, as the use of a threshold introduces discontinuities in the optimization by truncating the derivative chain.  
This truncation disrupts gradient flow, preventing the application of the standard backpropagation procedure.
Notably, even if sigmoid activations were directly used to evaluate a cost function based on the objective function, a threshold would still be inevitably applied.
In this hypothetical scenario, once effective solutions are identified, the network would repeatedly reinforce the same connections, limiting its exploration of alternative configurations and reducing its generalization capability.
Therefore, to effectively address this issue, the proposed training scheme incorporates the neural network's outputs by introducing a probabilistic evaluation of each $L$-dimensional vector, which is modeled using independent Bernoulli distributions.

Specifically, for each individual element $l_{k\ell}$ of the output vector, the corresponding random variable is modeled as $\mathcal{A}_{k\ell} \sim \Bernoulli(l_{k\ell})$, effectively transforming $l_{k\ell}$ into the probability that the $\ell$-th AP participates in the $k$-th UE's cluster.
By sampling from multiple Bernoulli distributions, we generate $K$ vectors, where each vector $\pmb{a}_k = [a_{k1}, \dots, a_{kL}]^\top \in \{0,1\}^L$ represents the link activations that are related to the $k$-th UE.
As is typical, this sampling process utilizes a uniform random variable $\mathcal{U}_{k\ell}$ over $[0,1]$ to obtain
\begin{equation*}
    a_{k\ell} = \begin{cases}
        1, &\text{if } \mathcal{U}_{k\ell} < l_{k\ell}\\
        0, &\text{otherwise.}
    \end{cases}
\end{equation*}

The learning process is conducted in mini-batches of size $R$, where each mini-batch is generated by simulating multiple independent channel realizations.
These realizations account for variations in large-scale fading coefficients due to shadowing effects, which depend on the random spatial distribution of UE positions within the designated reference area. 
By incorporating diverse channel conditions in each training epoch, the model learns to generalize across a broad range of network scenarios, improving its robustness.
Formally, for each \mbox{UE-AP} pair, $R$ large-scale fading coefficients are independently generated per training epoch. 
Consequently, the activation matrix corresponding to the $r$-th realization is defined as:
\begin{equation*}
    \mathbf{A}_r = \left[\left(\pmb{a}_1^{(r)}\right)^\top, \dots,\left(\pmb{a}_K^{(r)}\right)^\top \right]^\top
\end{equation*}
where $\pmb{a}_k^{(r)} \in \{0,1\}^L$ represents the link activation vector for the $r$-th realization, reflecting the state of the wireless connections between the $k$-th UE and the set of APs.
Each $\mathbf{A}_r$ determines the sets $\mathcal{L}_1, \dots, \mathcal{L}_K$ for the $r$-th realization, which serve as input to the module responsible for computing the selected objective function $\psi(\cdot)$. 
Once evaluated, the gradient of this function with respect to the activation matrix, $\nabla_{\mathbf{A}_r} \psi(\cdot)$, must be determined for each realization. 
To formalize the subsequent analysis, let us define the network output matrix 
\begin{equation}
    \mathbf{L}_r = \left[\left(\pmb{l}_1^{(r)}\right)^\top, \dots,\left(\pmb{l}_K^{(r)}\right)^\top \right]^\top
\end{equation}
where $\pmb{l}_1^{(r)}$ represents the network-generated output for the $k$-th UE in realization $r$.
The loss function that guides the training process for each realization is then expressed as
\begin{equation}
    \mathcal{C}_r (\pmb{\theta}) = \mathbf{L}_r (\pmb{\theta}) \odot \nabla_{\mathbf{A}_r} \psi(\cdot)
\end{equation}
where $\pmb{\theta}$ denotes the vector of all trainable parameters. 
This formulation explicitly couples the network outputs with the objective function’s gradient, ensuring that the optimization process is properly guided.
The total cost function over a mini-batch of $R$ realizations is finally obtained by summing the individual losses:
\begin{equation} 
    \mathcal{C} (\pmb{\theta}) = \sum_{r=1}^{R} \mathcal{C}_r (\pmb{\theta}). 
\end{equation}
Finally, the gradient of $\mathcal{C} (\pmb{\theta})$ with respect to $\pmb{\theta}$ is computed via backpropagation through time (BPTT).
These gradients are subsequently employed to update the parameters $\pmb{\theta}$ according to the selected optimization algorithm, iterating over multiple training epochs until convergence criteria are met.
The entire process described is presented in algorithmic form through the pseudocode shown in Algorithm \ref{alg:training}. 

\begin{algorithm}[t!]
\caption{Training Phase Algorithm}\label{alg:training}
\KwData{$L$ random positions for the APs and $1000$ random locations for each of the $K$ UEs}
\KwResult{The learned parameters $\pmb{\theta}$}

\For{$e \gets 1$ \KwTo Number of training epochs}{
    \For{each random position of the $K$ UEs}{
        Generate a batch $R$ of gains $\beta_{k\ell}$\;
        \For{$r \gets 1$ \KwTo $R$}{
            Sort the AP masters\;
            Sort the UEs for each master\;
            $\mathbf{L}_r \gets$ network outputs matrix\;
            $\mathcal{A}_{k}^{(r)} \sim \Bernoulli(\pmb{l}_{k}^{(r)})$\;
            $\mathbf{A}_r \gets$ samples from $\mathcal{A}_{k}^{(r)}$\;
            Compute $\nabla_{\mathbf{A}_r} \psi(\cdot)$\;
            Compute $\mathcal{C}_r (\pmb{\theta})$\;
        }
        Compute $\nabla_{\pmb{\theta}} \sum_{r=1}^R \mathcal{C}_r (\pmb{\theta})$\;
        Apply gradient through the optimizer\;
    }
}
\end{algorithm}

\subsection{Scalable Solution for UEs and APs}
\label{sec:scalability}
While the previously discussed AP-UE association algorithm provides clear insights into the methodological choices inherent in the master-centric approach and the probabilistic training framework, it faces significant scalability challenges regarding the number of APs. 
Indeed, while the algorithm effectively manages varying numbers of UEs, its architecture becomes increasingly unmanageable as the number of APs grows. 
This is because each component of the network produces $L$ outputs based on an input vector that encompasses all large-scale fading coefficients between the evaluated UE and every AP in the system.
As the number of APs increases, both the input size to the BiLSTM cell and the output of the FC network approach infinity, limiting the algorithm's usability.

To achieve this kind of scalability, it is crucial to limit the connectivity domain of each master AP. 
Although the network can expand indefinitely, each user connected to a specific master AP is restricted to a predefined group of proximate APs based on their location within a larger sub-area. 
In this framework, each master AP initiates the process by exchanging information with its neighboring APs to gather data about the users they manage. 
This collaboration enables each master AP to independently construct its own BiLSTM network, simulating the behavior of adjacent APs. 
As a result, each master AP operates autonomously, facilitating an efficient parallel decision-making process. 
From the entire simulated network, each master AP extracts only the decisions relevant to its associated users while disregarding other simulated outcomes.
A key feature of this fully scalable approach is that each network functions independently of users outside its designated sub-area. 
Consequently, due to the random distribution of users, each network typically manages a varying number of users in different scenarios, a flexibility enabled by the network's scalability regarding the number of UEs.

While the training of individual networks follows the procedure outlined in Algorithm \ref{alg:training}, the independent simulation by each master AP necessitates effective coordination. 
A key difference is that the vector $\pmb{z}_k$ (representing samples from the Bernoulli distribution) is generated solely by the master AP of the $k$-th UE and subsequently shared with its neighboring group APs.
Once this exchange is complete, all APs resume independent and parallel computation, evaluating the objective function based on their local simulated network.
This process ensures that each master AP generates the necessary gradients for updating its weights while maintaining a distributed learning framework. 
Note that the intra-group communication phase requires only synchronization among APs, ensuring seamless parallelization without disruption.

\subsection{Solution suited for scenarios with pilot contamination}
\label{sec:pilotcontamination}
For scenarios affected by pilot contamination due to the reuse of a limited set of orthogonal pilots for channel estimation, we introduce a refined version of our AP-UE association algorithm. 
Since the network has knowledge of the pilot assigned to each UE, this information can be incorporated into the scheme by expanding the input space of the BiLSTM cells, enhancing the network’s ability to learn and generalize in the presence of this kind of interference.
To implement this, we construct a one-hot vector with zeros in all positions except for a single `one' at the index corresponding to the pilot assigned to the $k$-th UE:
\begin{equation}
    \pmb{o}_k = [\underbrace{0, \dots, 0, \overset{\substack{t_k \\ \downarrow}}{1}, 0, \dots, 0}_{\tau_p}]^\top
\end{equation}
This vector is concatenated with the current UE features serving as input to the BiLSTM cell, forming the updated input $\hat{\pmb{\xi}}_k = \pmb{\xi}_k || \pmb{o}_k = [\pmb{\xi}_k^\top, \pmb{o}_k^\top]^\top$ for the $k$-th UE.
Although this method slightly increases the dimensionality of the input, it supplies the network with extra contextual information, resulting in a more robust decision-making capability and improved performance in presence of pilot contamination.

\section{Simulation results}
\label{sec:simulations}
To validate the proposed AP-UE association algorithms, we present results from extensive numerical simulations. 
We first analyze the potential of the scalable approach concerning the number of UEs and subsequently demonstrate that similar outcomes can be achieved through the fully scalable approach that also accounts for the number of APs.

\subsection{Performance analysis of the UE-scalable solution}
For the initial simulation scenario, we consider an area of $700 \times 700$ \unit{\square\m} comprising $L=25$ APs, each equipped with $N=4$ antennas. 
The AP locations are generated by randomly displacing their positions relative to an virtual regularly spaced grid, introducing random variations of up to $50\%$ of the inter-AP distance.
Within the same area, it is assumed that $K=10$ single-antenna UEs are also present.
For each UE, a set of multiple uniformly distributed locations are generated. 
Specifically, $1000$ locations are designated for the training set, while an additional $200$ locations are reserved for the test set, which is used to generate the subsequent results.

In all scenarios, we assume that the channel gain can be expressed in \unit{\dB} as follows:
\begin{equation}
    \beta_{k\ell}^{[\mathrm{dB}]} = \mathrm{SF}^{[\mathrm{dB}]}_{k\ell} - \mathrm{PL}^{[\mathrm{dB}]}_{k\ell} 
\end{equation}
Here, $\mathrm{SF}_{k\ell}$ denotes shadow fading, and $\mathrm{PL}_{k\ell}$ represents geometric path loss, determined according to the 3GPP microcell model \cite[Table B.1.2.1-1]{36.814}, valid for frequencies between $2$ and $6$ \unit{\GHz}:
\begin{equation}
    \mathrm{PL}^{[\mathrm{dB}]}_{k\ell} = 36.7 \log_{10} d_{k\ell} + 22.7 + 26 \log_{10} f_c \; ,
\end{equation}
where $d_{k\ell}$ denotes the distance from UE $k$ to AP $\ell$ (in meters), and $f_c$ represents the carrier frequency (in GHz).
Shadow fading is assumed to follow a lognormal distribution, specifically $\mathrm{SF}^{[\mathrm{dB}]}_{k\ell} \sim \mathcal{N}(0, \sigma_{\mathrm{SF}}^2)$, with  $\sigma_{\mathrm{SF}} = 4$ \unit{\dB} \cite{36.814}.
To take into account the existing correlation of the shadow fading from different UEs to the same AP, the following exponential function is typically adopted
\begin{equation}
	\mathbb{E}\left\{\mathrm{SF}^{[\mathrm{dB}]}_{i\ell} \mathrm{SF}^{[\mathrm{dB}]}_{j\ell}\right\} = \sigma_{\mathrm{SF}}^2 2^\frac{-\Delta_{ij}}{\delta_\mathrm{SF}}
\end{equation}
where $\Delta_{ij}$ is the distance (in meters) between the UE $j$ and UE $i$, and $\delta_\mathrm{SF}$ is the so-called ``correlation length'', which depending on the environment is hereinafter set to \SI{9}{\m} \cite[Table B.1.2.2.1-4]{36.814}. 
In our simulations, as typically done, we set the pre-log factor without considering the data uplink phase (i.e., $\tau_u = 0$).
All other communication simulation parameters are outlined in Table~\ref{tab:param}. 
Notably, the noise is assumed to be uniform in both the uplink and downlink, such that $\sigma_\mathrm{ul}^2 = \sigma_\mathrm{dl}^2$.

\begin{table}[!b]
    \caption{Simulation parameters}
    \centering
    \begin{tabular}{r|l}
        \textbf{Parameter} & \textbf{Value} \\ \hline
        Bandwidth & $20$ MHz \\
        Carrier frequency & $2$ GHz \\
        Samples per block $\tau_c$ & $200$ \\
        Noise power $\sigma_\mathrm{ul}^2 = \sigma_\mathrm{dl}^2$ & $-94$ dBm \\
        Per-UE uplink power $\eta_k$ & $100$ mW \\
        Per-AP maximum downlink power $\rho_\mathrm{max}$ & $200$ mW \\
        Difference in height between AP and UE & $10$ m
    \end{tabular}
    \label{tab:param}
\end{table}

The results of the conducted simulations are compared against three baseline methods for AP-UE association, which represent commonly used heuristic approaches in this field and demonstrate scalability with respect to the number of UEs.
The first two baselines employ a straightforward strategy of associating each UE to the $m$ APs with the largest values of the large scale fading coefficients \cite{buzzi2024co};
the third baseline, as presented in \cite{Bjornson2020}, focuses on mitigating interference among pilots by strategically selecting the optimal channel for each available pilot across all APs. 

\begin{figure*}[!t]
    \centering
    \subfloat[][]{\includegraphics[width=.33\linewidth]{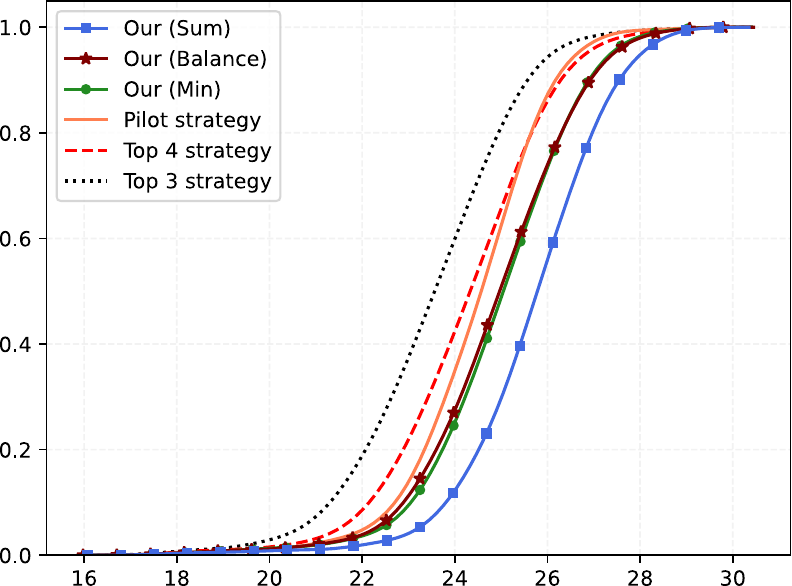}\label{fig:SE10}}
    \hfill
    \subfloat[][]{\includegraphics[width=.3125\linewidth]{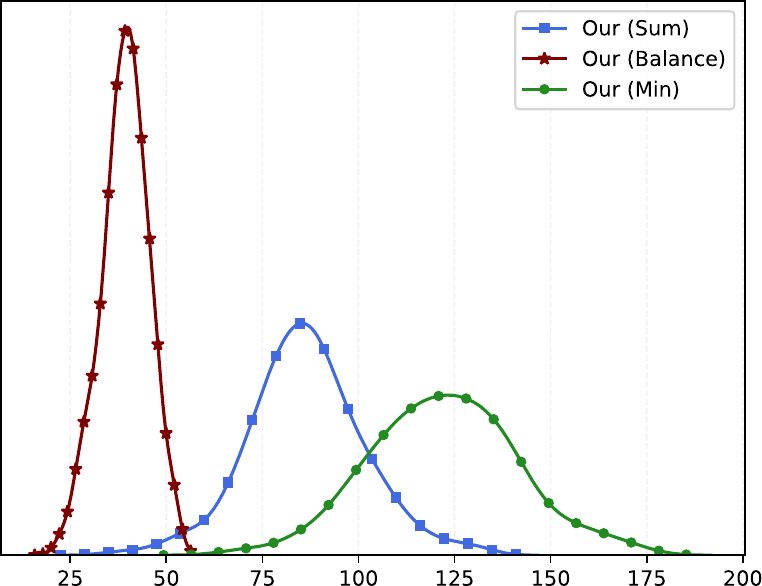}\label{fig:Conn10}}
    \hfill
    \subfloat[][]{\includegraphics[width=.33\linewidth]{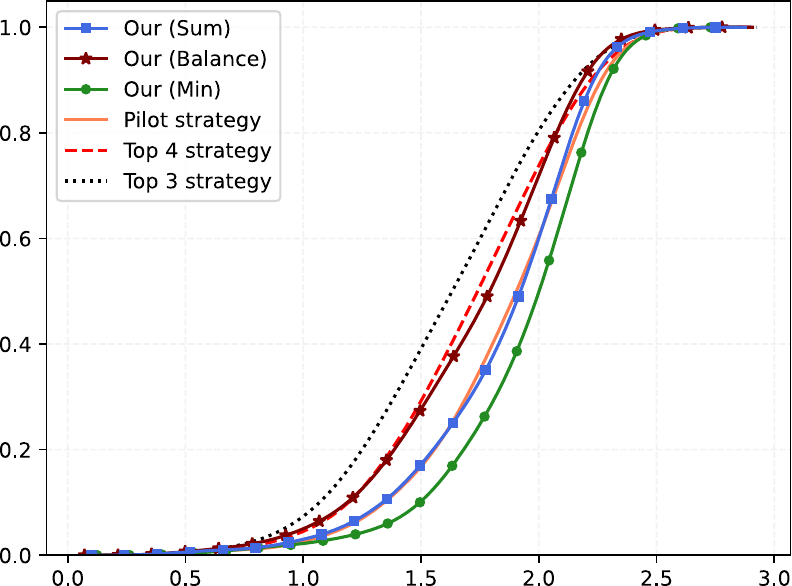}\label{fig:minimum}}
    \caption{Statistics for non-interfering pilot scenario: (a) CDFs of SE sums; (b) Distributions of activated connection counts; (c) CDFs of SE minima.}
    \label{fig:stats10}
\end{figure*}

Two simulation sets were conducted to evaluate the general applicability and effectiveness of the proposed method. 
The first scenario corresponds to the classical non-interference pilot scheme commonly found in the literature, in which the number of available orthogonal pilots matches the number of users in the system, thus assigning a unique pilot to each UE. 
Conversely, the second scenario assumes the availability of only $\tau_p = 4$ orthogonal pilots, resulting in pilot contamination in the system. In both cases, the network makes use of a hidden LSTM state size of $512$ and an FC network composed of three layers with $256$, $128$, and $25$ neurons, respectively. Each model developed for every reward function was trained over $200$ epochs, using the parameter $\lambda = 0.04$ for the $\psi_\mathrm{BALANCE}(\cdot)$ function.
Regardless of the operating scheme and sizes, the neural networks are consistently trained with a batch size of \mbox{$B = 64$} and using the ADAM optimization algorithm \cite{adam}, with a learning rate of $0.00001$ and default parameters.

\begin{table}[!b]
    \caption{Average Performance without Pilot Interference}
    \centering
    \begin{tabular}{lr|c|c|c} 
        && \textbf{\textit{SE sum}} & \textbf{\textit{SE min}} & \textbf{\textit{Connections}}\\\hline
        \multirow{3}{*}{\rotatebox[origin=c]{90}{\tiny OURS}} & \textbf{SUM} & $\pmb{25.71}$ & $1.85$ & $86.39$\\
        &\textbf{BALANCE} & $24.91$ & $1.73$ & $\pmb{39.23}$\\
        &\textbf{MIN} & $24.98$ & $\pmb{1.94}$ & $122.23$\rule[-0.5em]{0pt}{0pt}\\\hline
        \multirow{3}{*}{\rotatebox[origin=c]{90}{\tiny BASELINES}}\rule{0pt}{1em} & \textbf{Pilot strategy in} \cite{Bjornson2020} & $24.47$ & $1.85$ & $250$\\
        &\textbf{Top} $m=4$ \textbf{strategy} & $24.26$ & $1.72$ & $40$\\
        &\textbf{Top} $m=3$ \textbf{strategy} & $23.48$ & $1.62$ & $30$
    \end{tabular}
    \label{tab:perfor1}
\end{table}

\subsubsection{Non-Interfering Pilot Scenario}
Table \ref{tab:perfor1} presents the average results obtained on the test set in the absence of pilot contamination, with SE values expressed in \unit[per-mode = repeated-symbol]{\bit \per \s \per \Hz}. 
Specifically, it reports the average of the total SE, the average minimum SE per users, and the number of AP-UE connections for the strategy designed to maximize the three objective functions \eqref{eq:psi_sum}, \eqref{eq:psi_balance}, and \eqref{eq:psi_min}. 
These metrics are also provided for the three benchmark methods considered. 
The results confirm  that the proposed approach effectively aligns to each objective function, highlighting its ability to optimize diverse performance criteria.
Notably, all proposed strategies achieve higher average cumulative SE values compared to the baseline methods, with the $\psi_\mathrm{SUM}(\cdot)$ function outperforming the others. 
Additionally, it's crucial to observe that in this scenario, the pilot strategy attempts to associate each UE with every AP, setting an upper bound for the choose-the-top approach and indicating that performance degrades as the number of connections grows (i.e., if $m$ increases). 

%

The statistical data are presented in greater detail in Figure~\ref{fig:stats10}. Specifically, Figure~\ref{fig:SE10} illustrates the cumulative distribution functions (CDFs) for the sum of SE, emphasizing the superiority of the proposed approaches not only in terms of average performance. 
Figure~\ref{fig:Conn10} instead presents the distributions associated with the number of connections activated by the three proposed cost functions; while the baseline approaches remain fixed at the values indicated in Table \ref{tab:perfor1}.
This figure highlights that maximizing performance for either cumulative SE or minimum SE not only shifts the entire distribution towards higher average values but also widens its spread. 

\begin{figure}[!b]
    \centering
    \subfloat[][]{\includegraphics[width=.45\linewidth]{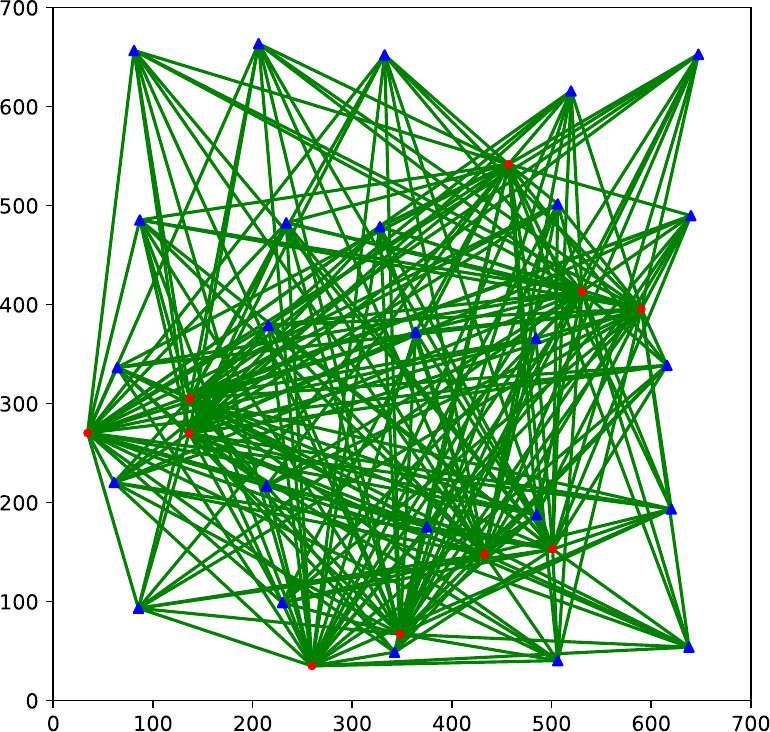}\label{fig:E1S}}
    \hfill
    \subfloat[][]{\includegraphics[width=.45\linewidth]{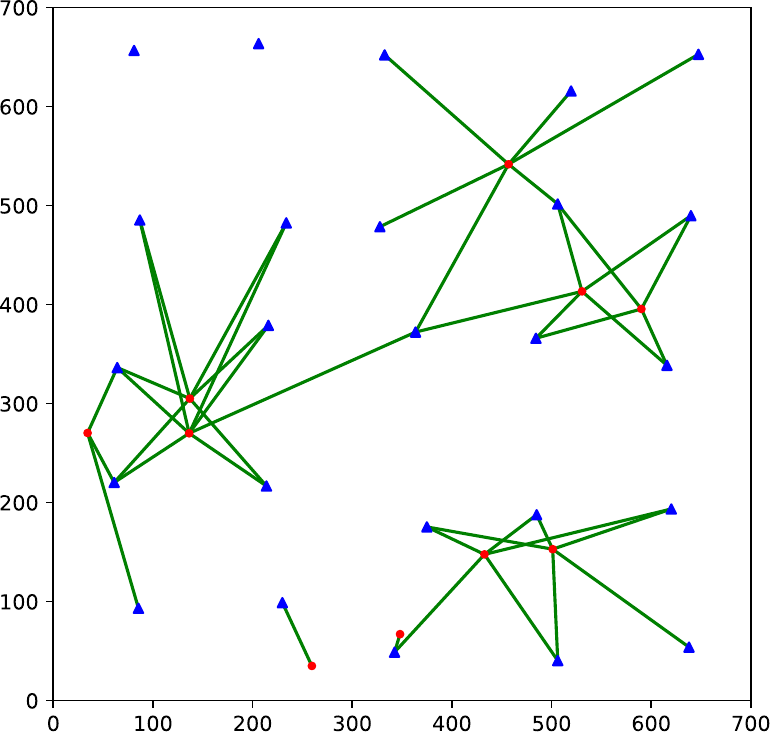}\label{fig:E1N}}
    \caption{Performance-matched example of active APs-UEs connections with $\tau_p=10$: (a) pilot strategy from \cite{Bjornson2020}; (b) $\psi_\mathrm{BALANCE}(\cdot)$ approach after training.}
    \label{fig:example1}
\end{figure}

A comprehensive analysis of these two graphs reveals that the $\psi_\mathrm{BALANCE}(\cdot)$ function significantly reduces the number of connections, achieving an average value comparable to the Top 4 strategy, while also delivering superior performance and better adapting to various interference scenarios.
In this evaluation, the comparison of active connections with equivalent performance, as illustrated in Figure~\ref{fig:example1}, is particularly noteworthy.
This figure, in fact, depicts a particular random scenario in which both the pilot strategy and $\psi_\mathrm{BALANCE}(\cdot)$ achieve a cumulative SE of approximately \SI[per-mode = repeated-symbol]{25.8}{\bit \per \s \per \Hz}, albeit with a significantly different number of connections ($250$ for the pilot strategy versus $43$ for our approach). 
Additionally, Figure~\ref{fig:E1N} underscores the proposed method's capability to activate single connections without compromising performance, even allowing some APs to remain unconnected.

A final consideration concerns the $\psi_\mathrm{MIN}(\cdot)$ approach, whose potential is not fully captured in Figure~\ref{fig:SE10} (focused on cumulative SE distribution).
This function is specifically designed to ensure equitable performance across UEs, increasing the minimum values even at the expense of the maximum ones.
Therefore, Figure~\ref{fig:minimum} provides a more insightful perspective, as it shows the cumulative distribution of the minimum SE for each test case, clearly demonstrating the approach's advantage in this aspect over the others.

\begin{figure*}[!t]
    \centering
    \subfloat[][]{\includegraphics[width=.33\linewidth]{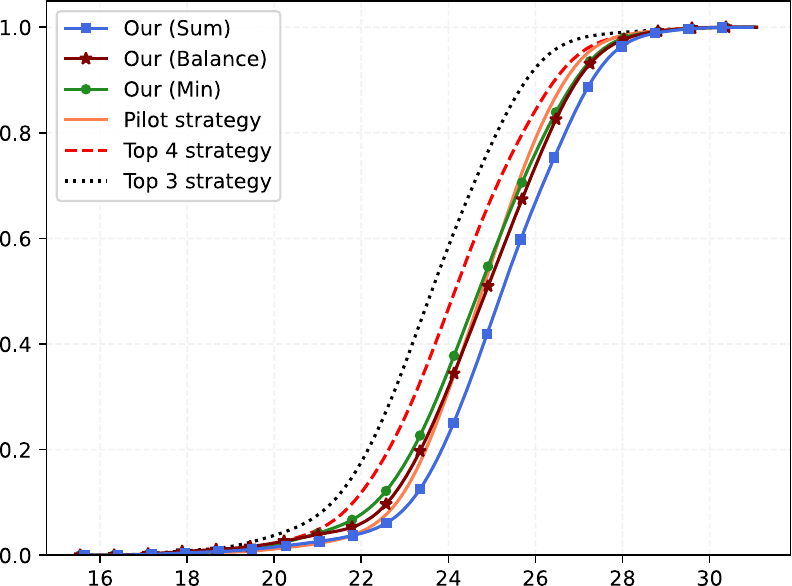}\label{fig:SE4}}
    \hfill
    \subfloat[][]{\includegraphics[width=.3145\linewidth]{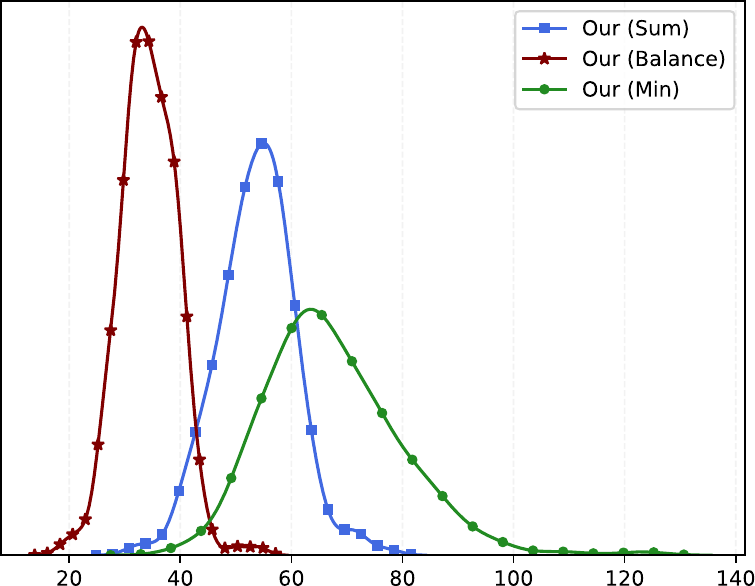}\label{fig:Conn4}}
    \hfill
    \subfloat[][]{\includegraphics[width=.33\linewidth]{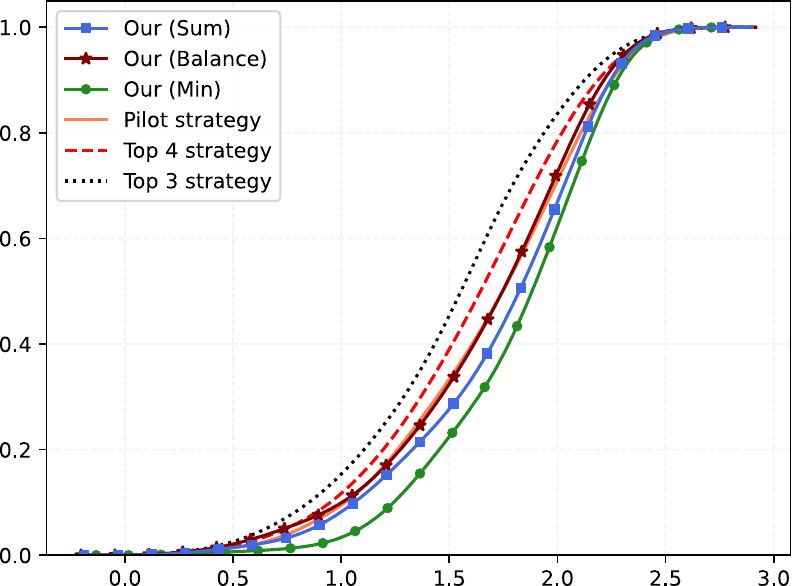}\label{fig:minimum4}}
    \caption{Statistics for high pilot interference scenario: (a) CDFs of SE sums; (b) Distributions of activated connection counts; (c) CDFs of SE minima.}
    \label{fig:stats4}
\end{figure*}

\begin{table}[!b]
    \caption{Average Performance with Pilot Interference}
    \centering
    \begin{tabular}{lr|c|c|c} 
        &&\textbf{\textit{SE sum}} & \textbf{\textit{SE min}} & \textbf{\textit{Connections}}\\\hline
        \multirow{3}{*}{\rotatebox[origin=c]{90}{\tiny OURS}} & \textbf{SUM} & $\pmb{25.18}$ & $1.73$ & $53.65$\\
        &\textbf{BALANCE} & $24.77$ & $1.67$ & $\pmb{34.29}$ \\
        &\textbf{MIN} & $24.63$ & $\pmb{1.82}$ & $67.40$\rule[-0.5em]{0pt}{0pt}\\\hline
        \multirow{3}{*}{\rotatebox[origin=c]{90}{\tiny BASELINES}}\rule{0pt}{1em} & \textbf{Pilot strategy in} \cite{Bjornson2020} & $24.73$ & $1.68$ & $100$\\
        &\textbf{Top} $m=4$ \textbf{strategy} & $24.13$ & $1.60$ & $40$\\
        &\textbf{Top} $m=3$ \textbf{strategy} & $23.57$ & $1.52$ & $30$
    \end{tabular}
    \label{tab:perfor2}
\end{table}

\subsubsection{High Pilot Interference Scenario}
For the scenario with pilot contamination, where $\tau_p=4$, the presented results are obtained by assuming the inputs described in Section \ref{sec:pilotcontamination}. 
The average performance on the test data is reported in Table~\ref{tab:perfor2}, reaffirming the effectiveness of the proposed methods in addressing each of the three distinct objective functions. 
The associated statistics, illustrated in Figure~\ref{fig:stats4}, demonstrate how the network mitigates pilot-induced interference by significantly reducing the average number of connections for all three functions compared to the interference-free case.
Notably, the required increase in connections for the $\psi_\mathrm{SUM}(\cdot)$ and $\psi_\mathrm{MIN}(\cdot)$ approaches (Figure~\ref{fig:Conn4}) follows a previously observed pattern, suggesting an intrinsic mechanism that increases the number of connections and broadens their distribution to achieve each objective. 
This trend is further supported by Figure~\ref{fig:minimum4}, which consistently highlights the distinctive performance of the $\psi_\mathrm{MIN}(\cdot)$ method in ensuring fairness.

\begin{figure}[!b]
    \centering
    \subfloat[][]{\includegraphics[width=.45\linewidth]{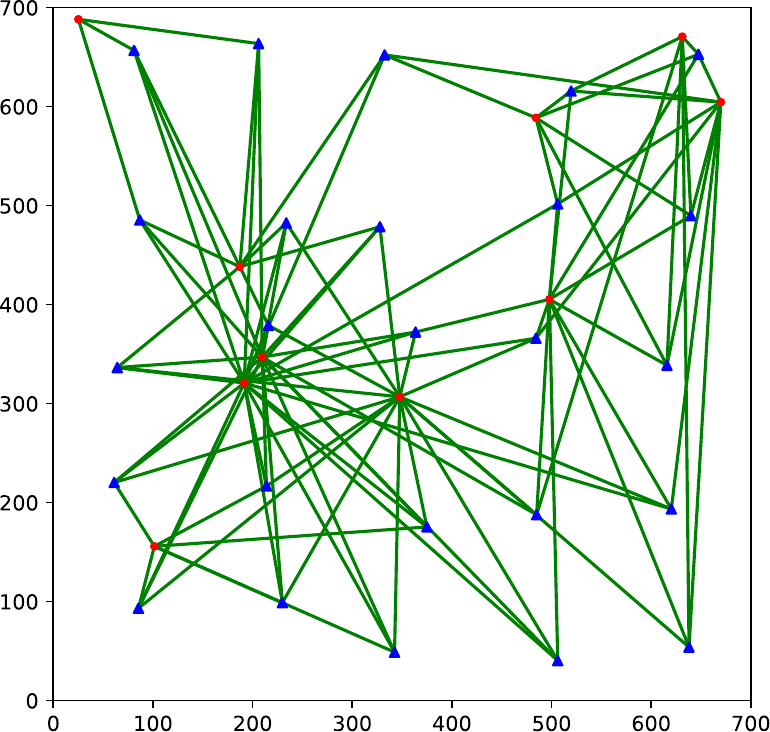}\label{fig:E2S}}
    \hfill
    \subfloat[][]{\includegraphics[width=.45\linewidth]{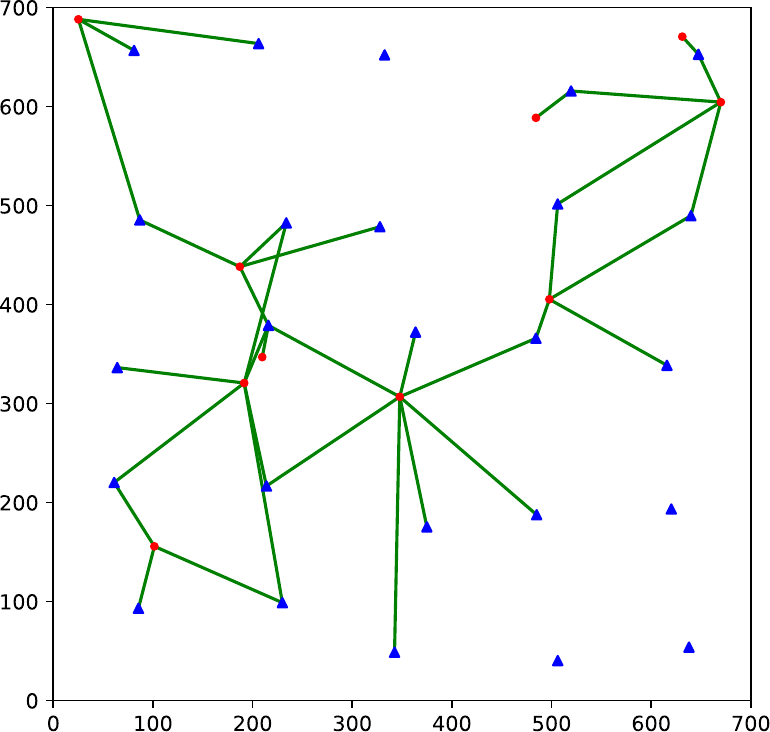}\label{fig:E2N}}
    \caption{Performance-matched example of active APs-UEs connections with $\tau_p=4$: (a) pilot strategy from \cite{Bjornson2020}; (b) $\psi_\mathrm{BALANCE}(\cdot)$ approach after training.}
    \label{fig:example2}
\end{figure}

In this case as well, it is insightful to compare the pilot strategy with the $\psi_\mathrm{BALANCE}(\cdot)$ function, particularly given their similar performance in terms of cumulative SE and minimum SE, as illustrated in Figures~\ref{fig:SE4} and \ref{fig:Conn4}. F
Figure~\ref{fig:example2} provides a direct comparison, focusing on a scenario where both approaches achieve a cumulative SE of approximately \SI[per-mode = repeated-symbol]{25.1}{\bit \per \s \per \Hz}.
Notably, even in this case, our function achieves this performance with approximately one-third of the active connections, leaving some APs entirely unconnected.

\subsection{Performance analysis of the fully scalable solution}
To evaluate the scalability of our approach in relation to the number of APs and validate its general applicability, we examine the method proposed in Section \ref{sec:scalability} in two different scenarios: an expanded one and a compressed one.
In the former scenario, we assess the same network configurations described earlier, assuming that each AP maintains a neighborhood range identical to that of the non-scalable approach previously discussed. 
This configuration is then adapted to an expanded setting where both the number of APs and UEs are increased.
Specifically, we consider an extended scenario with $49$ APs, arranged in the same regular grid pattern as before, serving a total of $20$ UEs.
It should also be noted that the AP-UE density per square kilometer remains approximately constant to ensure a fair comparison of results across different scales.
In the latter scenario, designed to evaluate the robustness of our approach in a highly constrained setting, we significantly limit each AP's awareness of its neighboring APs.
In particular, the network topology is restricted to a minimal neighborhood, where each AP is connected only to one adjacent AP in each cardinal direction.
Furthermore, within this reduced scenario, we also simplified the neural network architecture to test the adaptability of the model under conditions of lower complexity. 
In particular, the size of the hidden LSTM state was reduced to $256$, while the shared FC network was simplified to three layers with $128$, $64$, and $9$ neurons, respectively.

\begin{figure}[!b]
    \centering
    \subfloat[][]{\includegraphics[width=.45\linewidth]{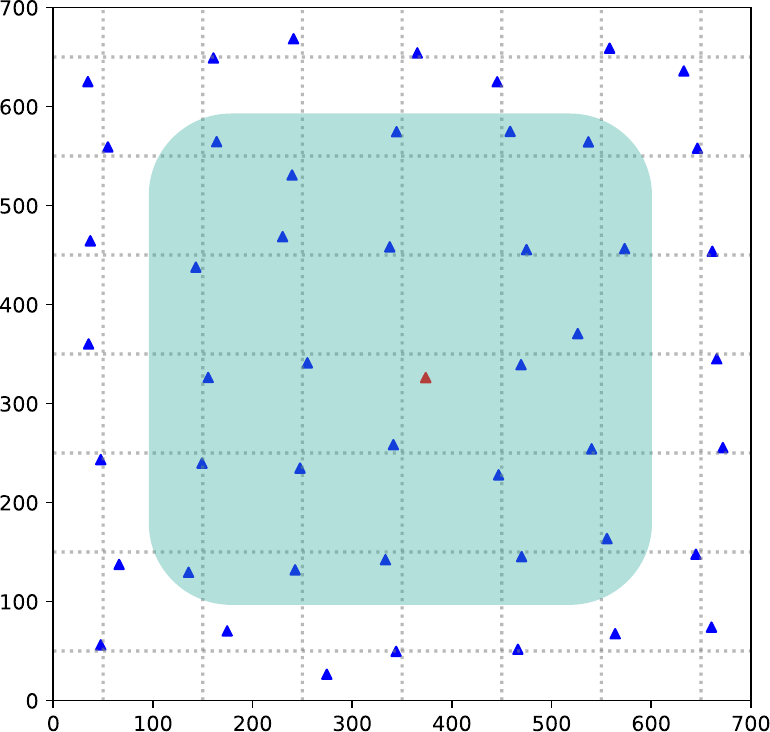}\label{fig:area2ext}}
    \hfill
    \subfloat[][]{\includegraphics[width=.45\linewidth]{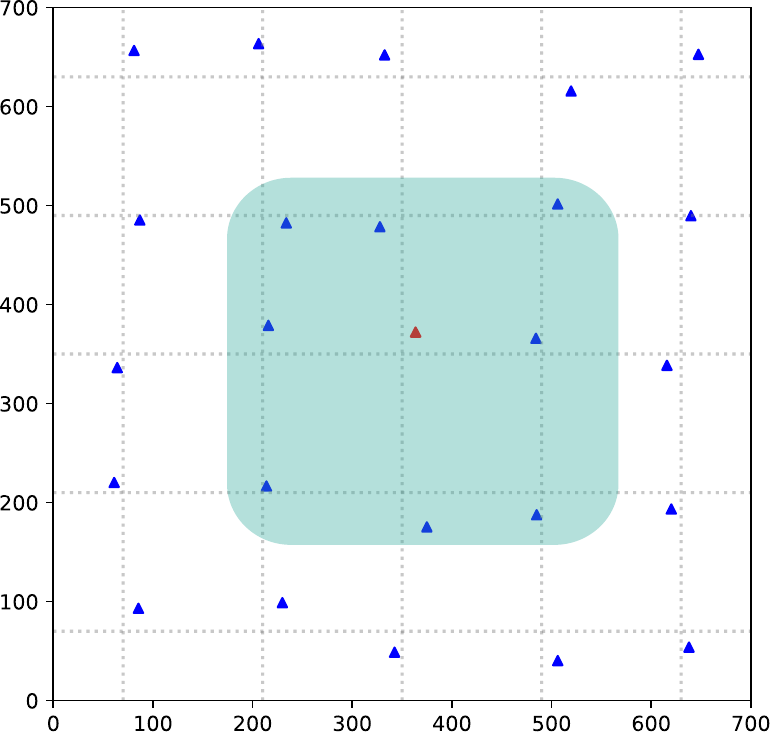}\label{fig:area2red}}
    \caption{Illustration of a network centered around the red master AP, showcasing all associated APs in both the extended (a) and compressed (b) scenarios.}
    \label{fig:area2}
\end{figure}

Figure \ref{fig:area2} illustrates these two scenarios, highlighting the APs involved in the output generation of the central red master AP. 
Remember that during the learning phase (exclusively in this phase) neighboring APs are required to share user information with the master AP, ensuring coordinated decision-making while adhering to the defined constraints.

In the subsequent analyses, for the sake of brevity, we will shift our focus exclusively on maximizing cumulative spectral efficiency, as this metric is the most sensitive to reductions in the application space, while assuming no interference between pilots.
Additionally, both scenarios were analyzed with a strong emphasis on maintaining a realistic perspective, which further exacerbated the inherent limitations. 
In particular, edge effects were taken into account, resulting in situations where corner APs could connect to a significantly reduced number of APs compared to those situated in the center. 
This focus on realism effectively rendered several of the output neurons inactive, demonstrating the robustness of the approach even in such challenging conditions.

\begin{figure}[!t]
    \centering
    \subfloat[][]{\includegraphics[width=0.8\linewidth]{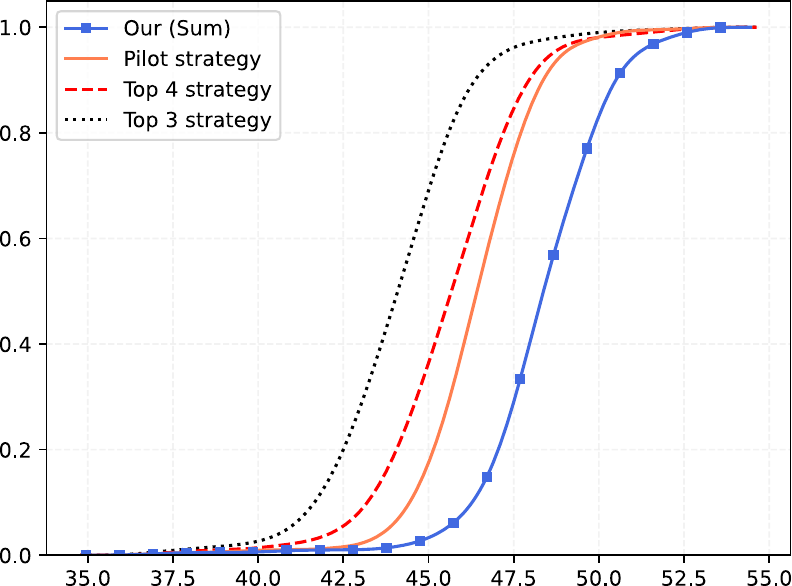}\label{fig:statsExt}}
    \hfill
    \subfloat[][]{\includegraphics[width=0.8\linewidth]{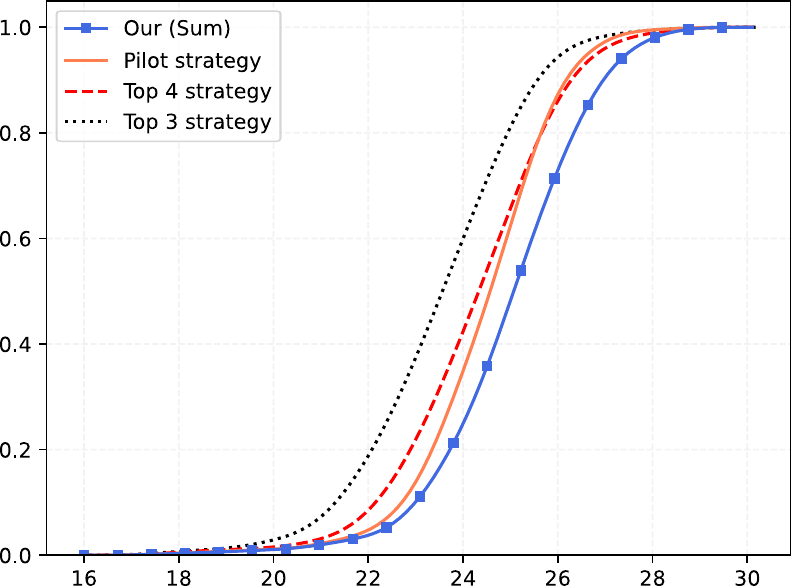}\label{fig:statsRed}}
    \caption{Statistics for both the extended (a) and compressed (b) scenarios.}
    \label{fig:statsSuper}
\end{figure}

The statistical results of the test simulations for the two scenarios are shown in Figure \ref{fig:statsSuper}.
Specifically, Figure \ref{fig:statsExt} highlights how the proposed architecture demonstrates a substantial performance gap compared to the baselines in the extended scenario. 
This is notable despite each AP operating and training its network within a reduced observation space relative to the global scenario.
It is worth noting that the spectral efficiency evaluated during the training phase is inherently biased, as it generally does not account for all the actual connections established. 
Additionally, each master AP independently produces its inference outputs without coordination with other APs.
Therefore, the displayed test results should be interpreted with this in mind, recalling that the tested UE configurations were not used during the training phase.

This improvement is also evident in the compressed scenario presented in Figure \ref{fig:statsRed}. 
In spite of the rigorous constraints imposed on the quantity of neighbors and the neural network’s structural dimensions, the proposed method demonstrates a distinct advantage. 
The compressed scenario has been intentionally crafted to assess the robustness of the proposed method. 
Even when subjected to stringent limitations on the reference space, which substantially impinge upon the network's potential, the method continues to retain its superiority. 
In contrast, the baseline methods remain unrestricted, thereby enabling connections throughout the entire space. 
This underscores the proposed approach's effectiveness and adaptability, as it surpasses conventional baselines under both relaxed and restrictive conditions. 
It illustrates not only exceptional performance in scenarios permitting greater flexibility but also resilience in environments that are highly constrained. 
These results substantiate the robustness and practical applicability of the proposed method across diverse levels of complexity and operational constraints.

\section{Conclusions}
This paper addresses the issue of AP-UE association in CF-mMIMO networks by proposing a DL-based algorithm that combines BiLSTM cells with a FC neural network, employing a master-centric approach alongside a probabilistic training methodology. 

This method demonstrates effective scalability with varying numbers of UEs without the need for re-training, achieving excellent performance and representing a robust solution even in mitigating pilot contamination.
Additionally, we have illustrated how our methodology can be applied to enhance scalability for both UEs and APs, observing excellent results in both expanded and compressed scenarios.
Numerical results illustrate the effectiveness of our proposed method, highlighting its superiority over traditional heuristics and indicating its potential for future wireless networks.

Interestingly, the proposed AP-UE association methodology can be adapted to optimize any objective function, suggesting various avenues for generalization and further research based on the findings of this study.

\bibliographystyle{ieeetr}
\bibliography{IEEEabrv, bibliography}

\end{document}